\documentclass[runningheads]{llncs}

\usepackage{graphicx}
\usepackage{amsmath,amssymb}
\usepackage{mathtools}
\usepackage{xspace}
\usepackage{booktabs} 
\usepackage{xcolor}
\usepackage{rotating}
\usepackage{colortbl}
\usepackage{multirow}
\usepackage{array}
\usepackage{hyperref}
\usepackage{url}
\newcommand{\al}{\textit{et al.}}

\renewcommand{\epsilon}{\varepsilon}
\newcommand{\R}{\mathbb{R}}

\renewcommand{\phi}{\varphi}

\usepackage{fullpage}

\title{Exploratory Landscape Analysis is Strongly Sensitive to the Sampling Strategy}
\titlerunning{ELA is Strongly Sensitive to the Sampling Strategy}

\author{
Quentin Renau\inst{1,2} \and
Carola Doerr\inst{3}\and
Johann Dreo\inst{1}\and
Benjamin Doerr\inst{2}
}
\authorrunning{Renau, Doerr, Dreo, Doerr}
\institute{
Thales Research \& Technology, Palaiseau France\\
\email{\{quentin.renau,johann.dreo\}@thalesgroup.com}
\and
Laboratoire d'Informatique (LIX), CNRS, \'Ecole Polytechnique, Institut Polytechnique de Paris, Palaiseau, France
\and
Sorbonne Université, CNRS, LIP6, Paris France\\
\email{carola.doerr@lip6.fr}
}

\begin{document}

\maketitle

\begin{abstract}
Exploratory landscape analysis (ELA) supports supervised learning approaches for automated algorithm selection and configuration by providing sets of features that quantify the most relevant characteristics of the optimization problem at hand. In black-box optimization, where an explicit problem representation is not available, the feature values need to be approximated from a small number of sample points. In practice, uniformly sampled random point sets and Latin hypercube constructions are commonly used sampling strategies. 

In this work, we analyze how the sampling method and the sample size influence the quality of the feature value approximations and how this quality impacts the accuracy of a standard classification task. While, not unexpectedly, increasing the number of sample points gives more robust estimates for the feature values, to our surprise we find that the feature value approximations for different sampling strategies do not converge to the same value. This implies that approximated feature values cannot be interpreted independently of the underlying sampling strategy. As our classification experiments show, this also implies that the feature approximations used for training a classifier must stem from the same sampling strategy as those used for the actual classification tasks.    

As a side result we show that classifiers trained with feature values approximated by Sobol' sequences achieve higher accuracy than any of the standard sampling techniques. This may indicate improvement potential for ELA-trained machine learning models.
\end{abstract}

\sloppy{
\section{Introduction} 
\label{sec:intro}

The impressive advances of machine learning (ML) techniques are currently shaking up literally every single scientific discipline, often in the function to support decisions previously requiring substantial expert knowledge by recommendations that are derived from automated data-processing techniques. Computer science is no exception to this, and an important application of ML is the selection and configuration of optimization heuristics~\cite{HutterKV19,kerschke_automated_2019,SmithMilesASsurvey}, where automated techniques have proven to yield tremendous efficiency gains in several classic optimization tasks, including SAT solving~\cite{SATzilla} and AI planning~\cite{vallati-ijcai15a}.  

In the context of numerical optimization, supervised learning approaches are particularly common~\cite{KerschkeT19,BelkhirDSS17,MunozASSurvey}. These methods often build on features developed in the context of \emph{fitness landscape analysis}~\cite{MalanE13surveyFLA,Pitzer12fitnesslandscape}, which aims at quantifying the characteristics of an optimization problem through a set of features. More precisely, a \emph{feature} maps a function (the optimization problem) $f:S \subseteq \R^d \to \R$ 
to a real number. Such a feature could measure, for example, the \emph{skewness} of $f$, its \emph{multi-modality}, or its similarity to a quadratic function.  

In practice, many numerical optimization problems are \emph{black-box problems}, i.e., they are not explicitly modeled but can be accessed only through the evaluation of \emph{samples} ${x \in S}$. Hyper-parameter optimization is a classical example for such an optimization task for which we lack an a priori functional description. 
In these cases, i.e., when $f$ is not explicitly given, the feature values need to be approximated from a set of $(x,f(x))$ pairs. The approximation of feature values through such samples is studied under the notion of \emph{exploratory landscape analysis} (ELA~\cite{mersmann_exploratory_2011}). ELA has been successfully applied, for example, in per-instance hyperparameter optimization~\cite{BelkhirDSS17} and in algorithm selection~\cite{KerschkeT19}.  
When applying ELA to a black-box optimization problem, the user needs to decide \textbf{how many samples to take} and \textbf{how to generate these}. 

When the functions are fast to evaluate, a typical recommendation is to use around $50d$ samples~\cite{kerschke_low-budget_2016}. For costly evaluations, in contrast, one has to resort to much fewer samples~\cite{BelkhirDSS16}. It is well known that the quality of the feature approximation depends on the sample size. Several works have investigated how the dispersion of the feature approximation decreases with increasing sample size, see~\cite{GallagherECJ19TrueValue,renau_expressiveness_2019} and references mentioned therein. The recommendation made in~\cite{kerschke_low-budget_2016} is meant as a compromise between a good accuracy of the feature value and the computational effort required to approximate it.   

Interestingly, the question which sampling strategy to use is much more widely open. In the context of ELA, the by far most commonly used strategies are uniform sampling (see, e.g.~\cite{morgan_sampling_2014,BelkhirDSS17}) and Latin Hypercube Sampling (LHS, see, e.g.,~\cite{kerschke_low-budget_2016,KerschkeT19}). These two strategies are also predominant in the broader ML context, although a few works discussing alternative sampling techniques exist (e.g.,~\cite{santner_design_2003}). A completely different approach, which we do not further investigate in this work, but which we mention for the sake of completeness, is to compute feature values from the search trajectory of an optimization heuristic. Examples for such approaches can be found in~\cite{DerbelLVAT19,JankovicD19}.  Note though, that such trajectory-based feature value approximations are only applicable when the user has the freedom to chose the samples from which the feature values are computed, a prerequisite not always met in practice. 

We share in this work the interesting observation that \textbf{the feature value approximations obtained from different sampling methods do not converge to the same values.} Put differently, the feature values are not absolute, but strongly depend on the distribution from which the $(x,f(x))$ pairs have been sampled. This finding is in sharp contrast to what seems to be common belief in the community. For example, it is argued in Saleem et al.~\cite[page 81]{GallagherECJ19TrueValue} that ``As $N$ [the sample size] $\to \infty$ the feature $\Phi_i$ approaches a true value''. We show in this work that no such ``true value'' exist: the feature values cannot be interpreted as stand-alone measures, but  only in the context of the samples from which they are approximated. 

Our observation has the undesirable side effect that machine-trained models achieve peak performance only when the sampling method applied to train the models was identical to the method used to approximate the feature values for the problem under consideration. Since the latter can often not be sampled arbitrarily (e.g., because we are forced to use existing evaluations), this implies that one might have to re-do a training ensemble from scratch. In application where we are free to chose the samples $(x,f(x))$, we need to ensure to store and share the points (or at least the distributions) that were used to approximate the feature values for the training data. Also, when using feature values to compare problems (see~\cite{GardenE14,DoerrDK19} for examples), one needs to pay particular attention that the differences in feature values are indeed caused by the function properties, and not by the sampling technique.

Given the sensitivity with respect to the sampling \emph{distribution}, one may worry that even the random number generator may have an impact on the feature value approximations. On a more positive note than the results  described above, we show that this is not the case. More precisely, we show that uniform sampling based on two very different random number generators, Mersenne Twister and RANDU, respectively, give comparable results.  

Another observation that we share with this paper is the fact that sampling strategies different from uniform and LHS sampling seem worth further investigation. More precisely, we show that classifiers trained with feature values that are approximated from samples generated by Sobol's low-discrepancy sequences perform particularly well on our benchmark problems. This challenges the state-of-the-art sampling routines used in ELA, and raises the question whether properties such as low discrepancy, good space-filling designs, or small stochastic dispersion correlate favorably with good performance in supervised learning.    

\textbf{Reproducibility:} The landscape data used for our analysis as well as several plots visualizing it are available at~\cite{dataPPSNpaper}.

\section{The Impact of Low Feature Robustness on Classification Accuracy}
\label{sec:task}
Instead of directly measuring and comparing the dispersion and the modes of the feature value approximation, we consider their impact on the accuracy in a simple classification task. We believe this approach offers a very concise way to demonstrate the effects that the different sampling strategies can have in the context of classical ML tasks. The classification task and its experimental setup is described in Sec.~\ref{sec:task2}.  
We then briefly comment on the distribution of the feature values (Sec.~\ref{sec:distri}), on the classifiers (Sec.~\ref{sec:classifier}), and on the sampling strategies (Sec.~\ref{sec:designs}). The impact of the low feature value robustness will then be discussed in Sec.~\ref{sec:accuracy1}, whereas all discussions related to the impact of the sampling strategy are deferred to Sec.~\ref{sec:strategy}.  

\subsection{Classification of BBOB Functions}
\label{sec:task2}

We consider the 24 BBOB functions that form the set of noiseless problems within the COCO (\underline{Co}mparing \underline{C}ontinuous \underline{O}ptimisers) platform~\cite{cocoplat}, a standard benchmark environment for numerical black-box optimization. For each BBOB problem we take the first instance of its 5-dimensional version. Each of these instances is a real-valued function $f:[-5,5]^5 \to \R$. The choice of dimension and instance are not further motivated, but are identical to those made in~\cite{renau_expressiveness_2019}, for better comparability.  

For the feature approximation, we sample for each of the 24 functions $f$ a number $n$ of random points $x^{(1)},\ldots,x^{(n)}$, and we evaluate their function values $f(x^{(1)}),\ldots,f(x^{(n)})$. The pairs $(x^{(i)},f(x^{(i)}))_{i=1}^n$ are then fed to the \emph{flacco} package~\cite{flacco}, which returns a vector of 46 features.\footnote{Note here that \emph{flacco} covers 343 features in total, which are grouped into 17 feature sets~\cite{kerschke_automated_2019}. However, following the discussion in~\cite{renau_expressiveness_2019} we only use 6 of these sets: dispersion (disp), information content (ic), nearest better clustering (nbc), meta model (ela\_meta), $y$-distribution (ela\_distr), and principal component analysis (pca).} 
We repeat this procedure 100 independent times, each time sampling from the same random distribution. This leaves us with 100 feature vectors per each function. From this set we use 50 uniformly chosen feature vectors (per function) for training a classifier that, given a previously unseen feature vector, shall output which of the 24 functions it is faced with. We test the classifier with all 50 feature vectors that were not chosen for the training, and we record the average classification accuracy, which we measure as the fraction of correctly attributed function labels. We apply 50 independent runs of this uniform random sub-sampling validation, i.e., we repeat the process of splitting the $24\times 100$ feature vectors into $24\times 50$ training instances and $24\times 50$ test instances 50 independent times. 

To study the effects of the sample size, we conduct the above-described experiment for three different values of $n$: $n=30$, $n=300$, and $n=5^5=3125$. 

The BBOB functions are designed to cover a broad spectrum of numerical problems found in practice. They are therefore meant to be quite different in nature. Visualizations of the functions provided in~\cite{bbob-functions} support this motive. We should therefore expect to see very good classification accuracy, even with non-tuned off-the-shelf classifiers. 

\subsection{Feature Value Distributions}
\label{sec:distri}

\begin{figure}[t]
    \centering
    \includegraphics[trim=0 10 0 0,clip,width=\textwidth]{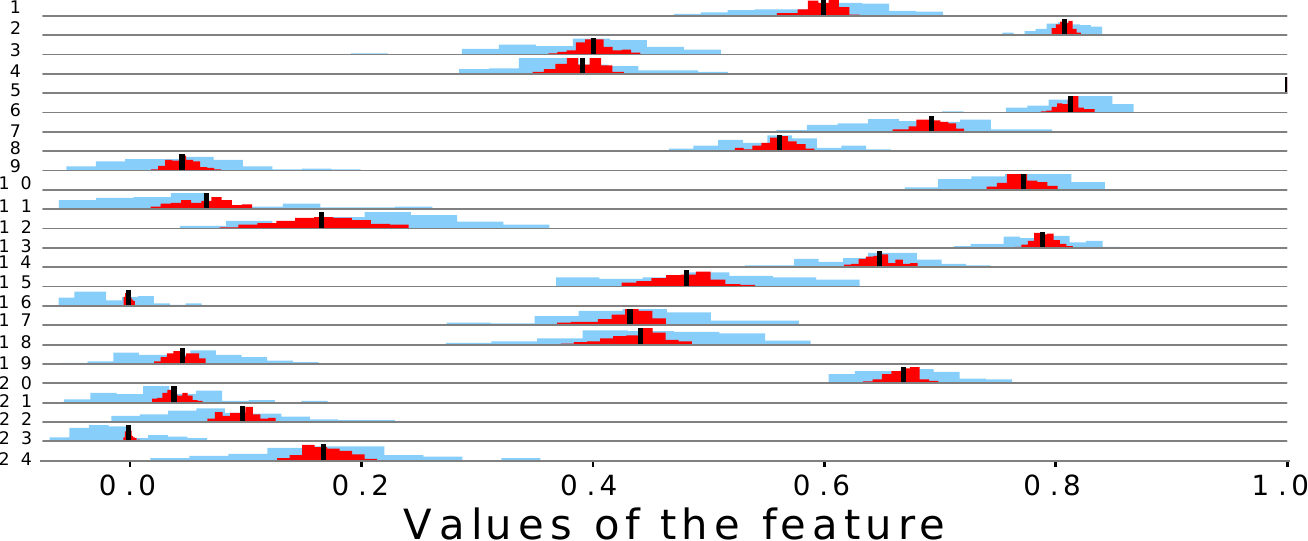}
    \caption{Distribution of the approximations for the ela\_meta.lin\_simple.adj\_r2 feature value, for 100 independently drawn LHS designs of $n=300$ (blue) and $n=3125$ (red) samples. Each row corresponds to one of the 24 BBOB functions. The black bars indicate the median value of the $n=3125$ data.}
    \label{fig:histogramms}
\end{figure}
  
Fig.~\ref{fig:histogramms} shows the distribution of the feature value approximations for one particular feature, which measures the adjusted fit to a linear model (observe that function 5 is correctly identified as a linear slope with an $R^2$ value of 1). Results are shown for $n=300$ (blue) and $n=3125$ (red) LHS samples.

We observe that the median values (black bars) of the single feature plotted in Fig.~\ref{fig:histogramms} are already quite diverse, i.e. -- taking a few exceptions aside -- they show fairly large pairwise distances. However, we also see that the dispersion of the approximated feature values is large enough to require additional features for proper classification. We also see that, in line with observations made in~\cite{GallagherECJ19TrueValue,renau_expressiveness_2019}, the dispersion of the approximations reduces with increasing sample size.   

\subsection{Classifiers: Decision Trees and KNN}
\label{sec:classifier}

All the classification experiments are made using the Python package \emph{scikit learn}~\cite[version~0.21.3]{scikit-learn}. Since we are not interested in our work to compare accuracy of different classifiers, but rather aim at understanding the sensitivity of the classification result with respect to the random feature value approximations, and since more sophisticated classifiers (in particular ensemble learning methods such as random forests) tend to require more computational overhead, 
we do not undertake any effort in optimizing the performance of these classifiers, and resort to default implementations of two common, but fairly different, classification techniques instead. Concretely, we use \textbf{$K$ Nearest Neighbors (KNN)} (we use $K = 5$) and \textbf{decision trees.} We decided to run the experiments with two different classifiers to analyze whether the effects observed for one method also occur with the other one. This should help us avoid reporting classifier-specific effects. For some selected results, we have performed a cross-validation with 5 independent runs of a random forest classifier, and found that -- while the overall classification results are better than for KNN and decision trees -- the structure of the main results (precisely, the results reported in Fig.~\ref{fig:heatmaps}) is very similar to that of the two classifiers discussed below. 

\subsection{Sampling Designs}
\label{sec:designs}

As mentioned previously, the two most commonly used sampling strategies in feature extraction, and more precisely in exploratory landscape analysis, are Latin Hypercube Sampling (LHS) and uniform random sampling. To analyze whether the sensitivity of the random feature value approximations depend on the strategy, we investigate a total number of five different sampling strategies, which we briefly summarize in this section. 

\paragraph{Uniform Sampling}
We compare uniform random sampling based on two different pseudo-random number generators: \\
-  \textbf{random:} We report under the name \emph{random} results for the Mersenne Twister~\cite{matsumoto_mersenne_1998} random number generator. 
    This generator is commonly used by several programming languages, including Python, C++, and R. It is widely considered to be a reliable generator. \\
- \textbf{RANDU:} we compare the results to those for the linear congruential number generator RANDU. 
    This generator is known to have several deficits such as an inherent bias that results in the numbers falling into parallel hyper-planes~\cite{knuth_art_1997}. We add this generator to investigate whether the quality of the random sampling has an influence on the feature value approximations and to understand (in Sec.~\ref{sec:strategy}) whether apart from the sampling strategy also the random number generator needs to be taken into account when transferring ELA-trained ML-models to new applications. 

\paragraph{Latin Hypercube Sampling (LHS)}
LHS~\cite{LHS} is a commonly used quasi-random method to generate sample points for computer experiments. 
In LHS, new points are sampled avoiding the coordinates of the previously sampled points. More precisely, the range of each coordinate is split into $n$ equally-sized intervals. From the resulting $n\times \ldots \times n$ grid the points are chosen in a way that each one-dimensional projection has exactly one point per interval.\\
- \textbf{LHS:} 
    Our first LHS designs  are those provided by the \emph{pyDOE} 
    Python package (version 0.3.8). We use the centered option, which takes the middle point of each selected cube as sample. \\
- \textbf{iLHS:} 
    The \emph{``improved''} LHS (\textbf{iLHS}) designs available in \emph{flacco}. This strategy builds on work of Beachofski et Grandhi~\cite{beachkofski_improved_nodate}. Essentially, it implements a greedy heuristic to choose the next points added to the design. At each step, it first samples a few random points, under the condition of not violating the Latin Hypercube design. From these candidates the algorithm chooses the one whose distance to its nearest neighbor is closest to the ideal distance $n/\sqrt[d]{n}$.

\paragraph{Sobol's low-discrepancy sequence}
We add to our investigation a third type of sampling strategies, the sequences suggested by \textbf{Sobol'} in~\cite{sobol_distribution_1967}. Sobol' sequences are known to have small \emph{star discrepancy}, a property that guarantees small approximation errors in several important numerical integration tasks. The interested reader is referred to~\cite{DickP10,Mat99} for an introduction to these important families of quasi-random sampling strategies. 

For our experiments we generate the Sobol' sequences from the Python package \emph{sobol\_seq} (version 0.1.2), 
with randomly chosen initial seeds.

\subsection{Classification Accuracy}
\label{sec:accuracy1}

\begin{figure}[t]
    \centering
    \includegraphics[width=\textwidth]{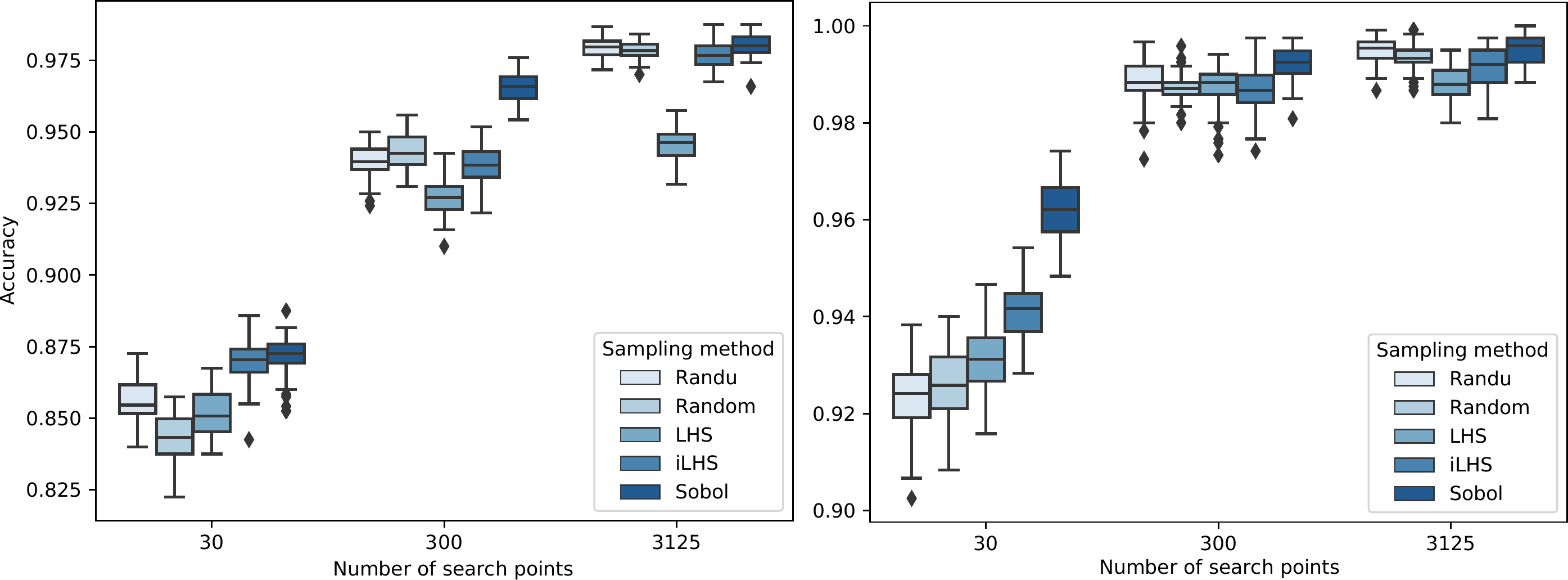}
    \caption{Classification accuracy by sampling strategy, sample size, and classifier (left=KNN, right=decision trees). Note the different scale of the $y$-axes.} 
    \label{fig:boxplots}
\end{figure}

Fig.~\ref{fig:boxplots} reports the distribution of the classification accuracy achieved by each of the five sampling strategies, when training and testing uses the same sampling strategy. The results on the left are for KNN classifiers, the ones on the right for decision trees. The absolute value of the medians can be inferred from Fig.~\ref{fig:heatmaps} (which we will discuss in Sec.~\ref{sec:strategy}). As expected, we see higher classification accuracy with increasing sample size. We also observe that the KNN results are slightly (but with statistical significance) worse than those of the decision trees. Recall, however, that this is not a focus of our search, and no fine-tuning was applied to the classification methods. Comparison between the two classifiers should therefore only be taken with great care. 

For KNN we nicely observe that the dispersion of the classification error reduces with increasing sample size. This aligns with the reduced variance of the feature value approximations discussed in Sec.~\ref{sec:distri}. For the decision tree classifier the dispersion of the classification accuracy reduces significantly from 30 to 300 samples, but then stagnates when increasing further to 3125 samples. 

No substantial differences between the two random number generators can be observed. For LHS, in contrast, the centered sampling method yields considerably worse classification accuracy than iLHS. 

Finally, we also observe that in each of the six reported (classifier, sample size) combinations the median and also the average (not plotted) classification accuracy of the Sobol' sampling strategy is largest, with box plots that are well separated from those of the other sampling strategies, in particular for $n\le 300$ samples. Kolmogorov-Smirnov tests confirm statistical significance in almost all cases. We omit a detailed discussion, for reasons of space. 

The good performance of Sobol' sampling suggests to question the state of the art in feature extraction, which considers uniform and LHS designs as default. Interestingly, our literature research revealed that Sobol' points were already recommended in the book of Santner \al~\cite{santner_design_2003}. It is stated there that Sobol' sequences may enjoy less popularity in ML because of their slightly more involved generation. Santner \al~therefore recommend LHS designs as fall-back option for large sample sizes. Our data, however, does not support this suggestion, and the (very small) advantages of the random sampling strategies over iLHS are indeed statistically significant.

\section{The Sampling Strategy Matters}
\label{sec:strategy}

Following the discussion above, it seems plausible to believe that the differences in classification accuracy is mostly caused by the dispersion of the feature value approximations. However, while this is true when we compare results for different sample sizes, we will show in this section that dispersion is not the main driver for differences between the tested sampling strategies.  

\begin{figure}[t]
    \centering
    \includegraphics[width=\textwidth]{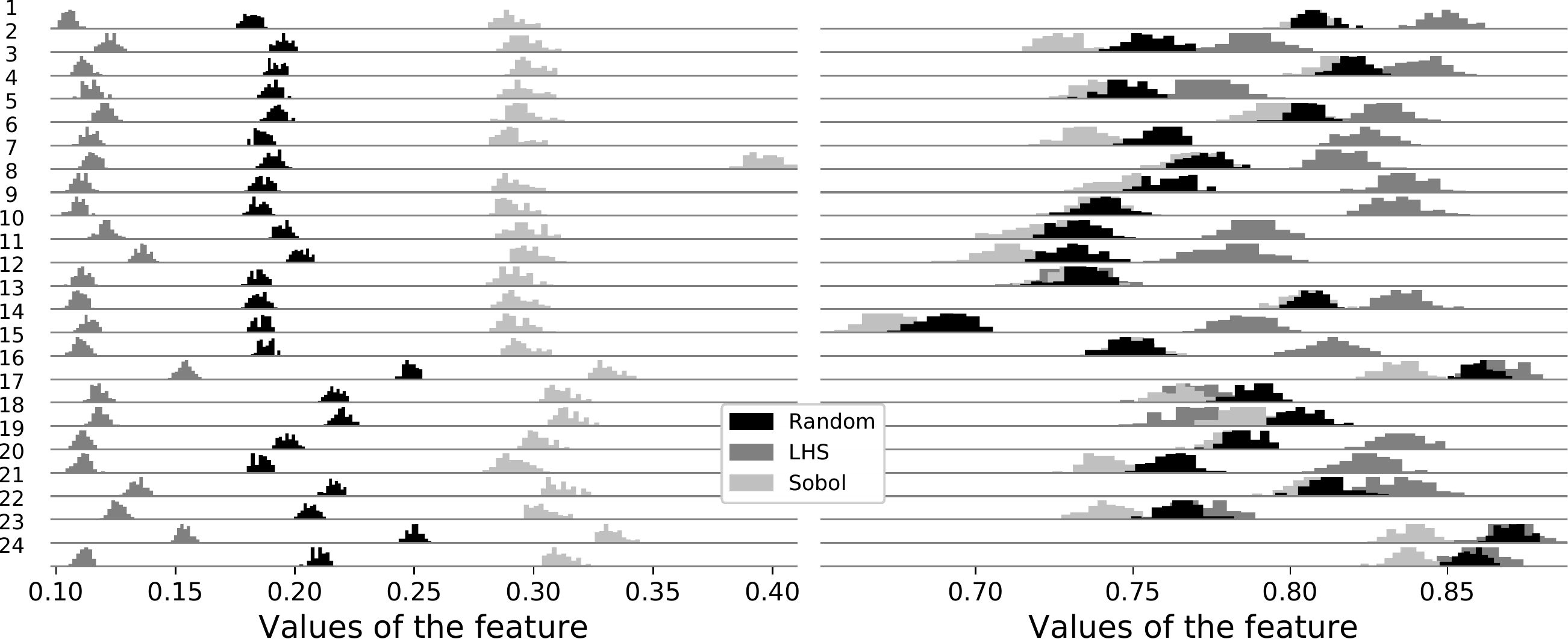}
    \caption{Distribution of feature value approximations for the nbc.dist\_ratio.coeff\_var feature (left) and ic.h\_max feature (right). 
    Results are for 100 independent evaluations of $n=3125$ samples generated by LHS, random, and Sobol' generators, respectively.}
    \label{fig:3strategies}
\end{figure}

Fig.~\ref{fig:3strategies} plots the distribution of feature value approximations for two of our 46 features. It illustrates an effect which came as a big surprise to us. Whereas features are typically considered to have an \emph{absolute} value (see the examples mentioned in the introduction), we observe here that the results very strongly depend on the sampling strategy. For the feature values displayed on the left, not only do the distributions have different medians and means, but they are even non-overlapping. This behavior is consistent for the different sample sizes (not plotted here). While this chart on the left certainly displays an extreme case, the same effect of convergence against different feature values can be observed for a large number of features (but not always for all functions or all different sampling strategies), as we can observe in the right chart of Fig.~\ref{fig:3strategies}. The latter also squashes hopes for simple translation of feature values from one sampling strategy to another one: looking at functions 10 and 12, for example, we see that random and Sobol' sampling yield similar feature values for both functions, whereas those approximated by LHS sampling are much larger for f10 as for f12.  
\textbf{We thus observe that the interpretation of a feature value cannot be carried out without knowledge of the sampling strategy.} 

\begin{figure}[t]
    \centering
    \includegraphics[width=\textwidth]{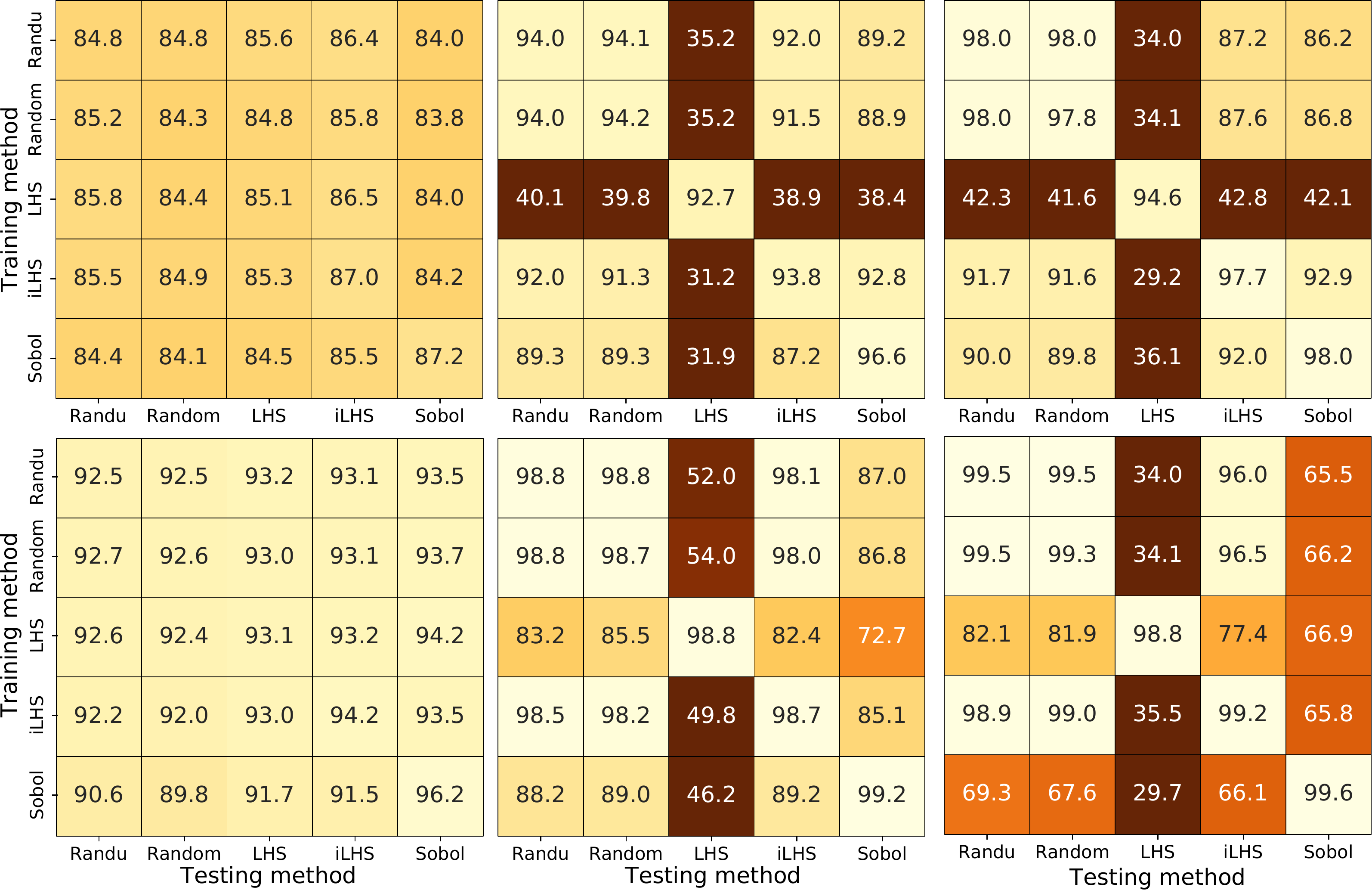}
    \caption{Heatmaps of median classification accuracy for KNN (top) and decision trees (bottom), for feature values approximated by 30 search points (left), 300 search points (middle), and 3125 search points (right), respectively.}
    \label{fig:heatmaps}
\end{figure}

We investigate the impact of the strategy-dependent feature values by performing the following classification task. We use the same feature values as generated for the results reported in Sec.~\ref{sec:task}, but we now train the classifiers with the feature value approximations obtained from one sampling strategy, and we track the classification accuracy when tested with feature value approximations obtained by one of the other strategies. Apart from this twist, the experimental routine is the same as the one described in Sec.~\ref{sec:task2}.

The heatmaps in Fig.~\ref{fig:heatmaps} display the median classification accuracy of the 25 possible combinations of training and testing sampling strategies. We show results for all three sample sizes, $n=30$ (left), $n=300$ (middle), and $n=3125$ (right). Rows correspond to the training strategy, the columns to the test strategy; the diagonals therefore correspond to the data previously discussed in Sec.~\ref{sec:accuracy1}. KNN data is shown in the top, those for decision trees on the bottom.

For sample size $n=300$ and $n=3\,125$ the best or close-to-best classification accuracy is achieved when the sampling strategy for the testing instances is identical to that of the training instances. This is independent of the classifier. Interestingly, this observation does not apply to the case with $n=30$ samples, where, e.g., the KNN classifiers trained with LHS data achieve better accuracy with iLHS feature approximations (86.5\%) than with LHS approximations (85.1\%). The same holds for the classifiers trained with data from the random sampling strategy (for both random number generators). The differences between the different training and test data combinations, however, are rather small in these cases. In addition, the dispersion of the classification accuracies are relatively large for $n=30$ samples, with ranges that are very similar to those plotted in Fig.~\ref{fig:boxplots}. We also recall that the overall classification accuracy, in light of the high diversity of the 24 BBOB functions, is not as good as it may seem at the first glance. 

We also observe that, for $n=30$, the KNN classifiers (except for the Sobol'-trained ones) perform best when tested with iLHS test samples, whereas for decision trees we see better results with Sobol' test data. This however, applies only to the case $n=30$, as we shall discuss in the following. 

Moving on to the cases $n\ge 300$, we observe that -- in line with the observation made in Fig.~\ref{fig:boxplots} -- the average classification accuracy increases significantly, to values well above 90\%, with a few notable exceptions: The poor accuracy of LHS both as test and as training instances stands out, but is consistent for both classifiers, and both sample sizes $n=300$ and $n=3\,125$. Albeit not as bad, the Sobol'-approximated feature values also lead to comparatively poor performance on almost all classifiers not trained with Sobol'-approximations (an exception are the iLHS-trained KNN classifiers using $n=300$ samples). Consistent to this, the Sobol'-trained classifiers have low classification accuracy when tested with feature values from the other four strategies. While this effect is most noticeable for the decision tree classifiers, it also applies to KNN. 
A closer inspection of the feature value approximations reveals that those for iLHS, random sampling, and Randu are much more alike to each other than to the LHS or Sobol' features. For $947=43\%$ of all $24\times 46$ (function, feature) pairs, the median of the LHS feature values with $n=3125$ samples is either smaller or larger than that of the other strategies. For Sobol' points, this value is $725=33\%$. Of course, this just gives a first impression. Plots similar to Figure~\ref{fig:3strategies} provide much more details; they are available for all features at~\cite{dataPPSNpaper}. A thorough investigation into \emph{why} these differences exist forms an important next step for our research, cf. Sec.~\ref{sec:conclusions}. 

Given that we use the centered option for the LHS strategy (see Sec.~\ref{sec:designs}), one might be tempted to think that the LHS-approximations are more concentrated than those of the other sampling strategies. This, however, cannot be confirmed by our data: the dispersion of the LHS approximations is comparable to that of the other strategies.  
 
 Finally, we observe that the two random strategies show high inter-strategy classification accuracy. Their feature approximations work furthermore quite well with classifiers trained on iLHS data. However, while all of the results reported above also apply to average (instead of median) classification accuracy, the average classification error of the iLHS-trained KNN-classifiers is considerably worse for (Mersenne-Twister) random feature value approximations than for those obtained from Randu (91.4\% vs. 94.0\% accuracy for $n=300$ and 91.7\% vs. 98\% for $n=3125$ samples).  
 
 \section{Confusion Matrices}
 \label{sec:confusion} 

\begin{figure}[t]
    \centering
    \includegraphics[width=\textwidth]{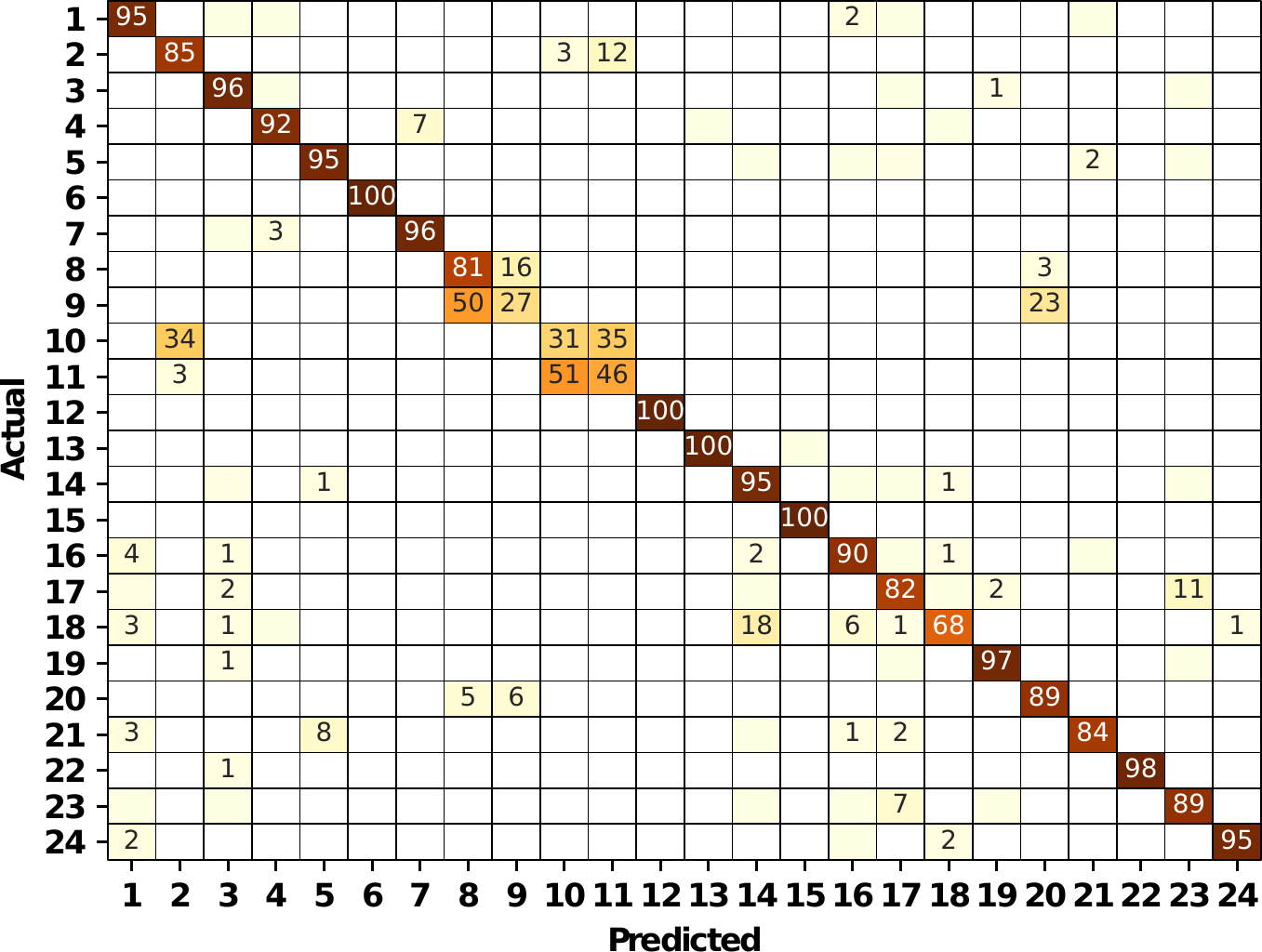}
   \caption{Average classification result across 50 independently iLHS-trained KNN classifiers, each tested with 50 Sobol' feature value approximations using $n=30$ evaluations. Numbers are provided for $>1\%$ probabilities only.
   }
    \label{fig:confusion}
\end{figure}

The results reported in the previous sections were all aggregated over the 24 functions from the BBOB benchmark set. In Fig.~\ref{fig:confusion} we analyze which functions are misclassified most frequently, and by which functions they are confused with. The matrix shows results for the 50 KNN classifiers trained with iLHS feature approximations and tested with Sobol' data (50 tests per classifier), for $n=30$ sample points. We recall from Fig.~\ref{fig:heatmaps} that the median classification accuracy of this combination is 84.2\%. This is also the average accuracy. 

Most functions are correctly classified with probability at least 80\%. For twelve functions we observe at least 95\% accuracy. Only four functions (9-11, 18) are misclassified  with probability $\ge30\%$, and those are typically confused by the same one or two other functions. Function 2, for example, is misclassified as function 11 in 12\% of the tests. 

We do not show the confusion matrices for the other $3\times 25$ cases, but note that -- overall -- the patterns are quite consistent across all KNN classifiers. Naturally, the concentration on the diagonal increases with larger sample sizes. We also see a higher concentration for the mis-classifications as well. For example, in the same iLHS-Sobol' setting as above with $n=3125$ samples 15 functions have accuracy $\ge 95\%$, and only five function pairs with mis-classification rate $\ge 5\%$ are observed. Four of these occur with probability $\le 8\%$. One mis-classification stands out: function 9 is classified as function 20 in 93\% of the cases.

For decision trees, the structure of the confusion matrices is similar to those of KNN for $n=30$ samples. For $n=3125$ samples, however, the mis-classifications are much more scattered across several function pairs.

Without going further into details, we note that these results can be used to understand deficits of the benchmark sets (frequent confusion of two functions could indicate some problems are quite alike), of the selected features (if they do not capture major differences between two structurally different functions), and the classifiers (e.g., the scattered confusion matrix of the decision trees for $n=3125$ samples).

\section{Conclusions}
\label{sec:conclusions}

We have analyzed the impact of the stochasticity inherent to feature value approximations on the use of exploratory landscape analysis in classic ML tasks. Our key findings are the following. 

\textbf{(1) ELA features are not absolute, but should be interpreted only in the context of the sampling strategy.} As an important consequence of this observation, we derive the recommendation that the sampling strategy of the training data should match the sampling strategy of the test data. Note that this also implies more data needs to be shared to obtain reproducible and/or high quality results.

\textbf{(2) The good results achieved by the classifiers trained with Sobol' samples suggests to revive a recommendation previously made by Santner \al~\cite{santner_design_2003}, and to further investigate this sampling strategy in the context of other feature extraction tasks, i.e., beyond applications in exploratory landscape analysis.} 
In this context, it would also be worthwhile to study other low-discrepancy constructions, which are recently gaining interest in the broader ML context, e.g., in the context of \emph{one-shot optimization} (the task of optimizing a black-box problem through the best of $n$ parallel samples, see~\cite{Teytaud19} and references therein). Whether good performance in one-shot optimization correlates with a good approximation of feature values forms another interesting avenue for future work.  

While we have focused in this work on classification accuracy only, we are also planning on a more detailed analysis of the feature approximations themselves. In particular, we aim at understanding a functional relationship between the sampling strategies and their feature value approximations. This shall help us identify correction methods that translate values obtained from one sampling strategy to another. This, ultimately, may help us by-pass the need for sample-specific training.

We also believe that the confusion matrices such as the one in Fig.~\ref{fig:confusion} should be explored further, to understand which BBOB instances are more alike than others. Such information can be useful for instance selection and generation.   

\vspace{1ex}
{\footnotesize{
\textbf{Acknowledgments.} 
This research benefited from the support of the FMJH Program PGMO and from the support of EDF and Thales. 
}}

}

\end{document}